\documentclass{article}
\pdfoutput=1
\usepackage{graphicx}
\usepackage{multirow}
\usepackage{url}
\usepackage{geometry}
\usepackage{tikz}
\usepackage{pgfplots}
\pgfplotsset{compat=1.18} 
\usepgfplotslibrary{fillbetween}
\usepackage{titling}
\usepackage{authblk}
\usepackage[backend=biber,sorting=none]{biblatex}
\date{\vspace{-5ex}}

\title{Visual Car Brand Classification by Implementing a Synthetic Image Dataset Creation Pipeline}

\author[1]{Jan Lippemeier}
\author[2]{Stefanie Hittmeyer}
\author[2]{Oliver Niehörster}
\author[3]{Markus Lange-Hegermann}
\affil[1]{TH OWL University of Applied Sciences and Arts}
\affil[2]{Fraunhofer IOSB-INA}
\affil[3]{Institute for Industrial Information Technology (inIT)}

\begin{document}

\maketitle

\begin{abstract}
Recent advancements in machine learning, particularly in deep learning and object detection, have significantly improved performance in various tasks, including image classification and synthesis. However, challenges persist, particularly in acquiring labeled data that accurately represents specific use cases.
In this work, we propose an automatic pipeline for generating synthetic image datasets using Stable Diffusion, an image synthesis model capable of producing highly realistic images. We leverage YOLOv8 for automatic bounding box detection and quality assessment of synthesized images.
Our contributions include demonstrating the feasibility of training image classifiers solely on synthetic data, automating the image generation pipeline, and describing the computational requirements for our approach. We evaluate the usability of different modes of Stable Diffusion and achieve a classification accuracy of 75\%. 
\end{abstract}
\section{Introduction}

In the last twelve years advancements in machine learning, deep learning and object detection achieved remarkable results in performance. 
The deep learning revolution in image classification started with the publication of AlexNet \cite{Alexnet} in 2012. 
Further performance enhancements have been achieved in the following years with models such as ResNet \cite{Resnet}. 
Further the subject of object detection achieved high results with models such as YOLO \cite{YOLO}.
The potential for synthetic data in identification and classification tasks has already been shown \cite{pub.1147357025, 10508671}.
Recent developments in image synthesis such as Stable Diffusion \cite{StableDiffusion} achieved a high-level performance on generating realistic images. Related works have already used synthetic images from Stable Diffusion as training data \cite{10244720, 10499667}. 

Transfer learning which uses pre-trained existing models is a common approach for solving image classification tasks \cite{TransferLearning, TransferLearningExample}. 
However, this approach requires large amounts of labeled data.
Existing publicly available datasets can only be successfully used for training if they actually resemble the use case.
Even if large amounts of data from the actual use case can be acquired, labeling this data remains time consuming and therefore expensive, as this is often a manual task.
Biases within the data present a challenge as there is a compromise to be made, either in the form of potentially keeping the bias, reducing the dataset size to balance classes or oversampling underrepresented classes. 
In most cases this leads to small available datasets and consequently overfitting.
Another common challenge is a low variance within the available data.

Solving computer vision tasks when only limited data is available, is a common major challenge in practice.
Limited datasets often lead to overfitted or poorly performing models, endangering the success of a project.
We face this challenge by illustrating an adaptable approach. For this approach we synthesize images on demand that are tailored to the respective use case. This leads us to posing our abstract main research question: Is it possible to use image synthesis in an automated manner to create suitable datasets for computer vision tasks with otherwise limited existing data?
We evaluate this general approach on a specific real-world application.

\begin{figure}[t]

    \centering

    \begin{tabular}{cccccccc}
    
     \includegraphics[width=0.11\textwidth,height=0.11\textwidth]{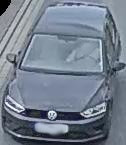} 
     &  \includegraphics[width=0.11\textwidth,height=0.11\textwidth]{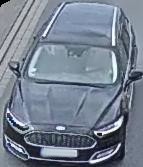}
     &  \includegraphics[width=0.11\textwidth,height=0.11\textwidth]{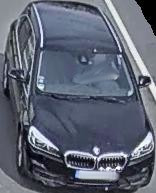}
     &  \includegraphics[width=0.11\textwidth,height=0.11\textwidth]{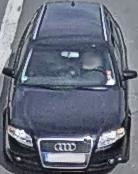}
     &  \includegraphics[width=0.11\textwidth,height=0.11\textwidth]{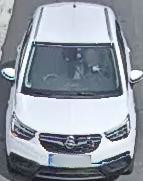}
     &  \includegraphics[width=0.11\textwidth,height=0.11\textwidth]{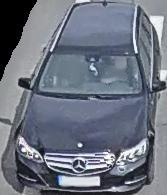}
     &  \includegraphics[width=0.11\textwidth,height=0.11\textwidth]{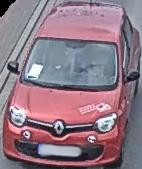}
     &  \includegraphics[width=0.11\textwidth,height=0.11\textwidth]{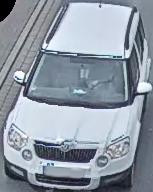} \\
     Volkswagen& Ford & BMW & Audi & Opel & Mercedes & Renault & Skoda
    \end{tabular}

    \caption{The selected brands we aim to classify. These eight brands occur the most in our recorded footage.}
    \label{fig:Selected Brands}
\end{figure}

Our real-world image classification task is to visually predict the brand of a car as an unequivocal visually determinable feature (see Figure \ref{fig:Selected Brands}). 
We recorded and labeled data that represents the German automotive traffic; this data was recorded by traffic cameras in Lemgo - a medium sized town in Germany.
However, we are limited by the traffic volume and the capacities for human labeling. 
Even if unlimited gathering of real labeled data from the German traffic was possible, the data would still include the biases of the real world. 
Filtering these out and labeling the images would still remain a time consuming task.
Existing related datasets such as the Stanford Car Dataset \cite{Stanford_Car_Dataset} tend to resemble the North American market.
Some car brands common in Germany are normally not even present within existing datasets.
With limited data from the actual application and no usable existing dataset this problem is a prime example for our main research question.

\setlength{\tabcolsep}{3pt}
\setlength{\tabcolsep}{6pt}

With the emergence of image synthesis models that create highly realistic images with correct proportions and details we propose the usage of synthetic images as training data for image classification tasks. 
In theory image synthesis models are a prompt-guided way to synthesize an image of a desired object.
Stable Diffusion is an open source model that allows for programmatic image synthesis \cite{StableDiffusion}. The creation of images therefore becomes a question of time and computing power.

We create an automatic pipeline for image dataset creation using Stable Diffusion as a tool to synthesize images. We are able to control the distribution of the generated images by controlling the distribution of the used prompts. By using YOLO we automatically determine the bounding boxes of a car inside a generated image. YOLO also allows us to estimate whether the synthetic image is suitable by giving a confidence score, the bounding box and the class of the detected object.

While we are able to avoid including biases in regards to distribution over classes by balancing out the generated images per manufacturer, image synthesis models may still include an inherent bias. 
If there are biases within the training data of the image synthesis model it might pass this bias on to the generated images. 
In regards to cars the data used in the training of Stable Diffusion could be unbalanced, for example, by favoring new over old, famous over unfamous, and popular over unpopular cars. 
Further it is not automatically confirmable whether a synthetic image actually matches the desired output encoded by the prompt.
Also the perspective, illumination, contrast and other photographic properties might differ from the actual task.
~\\
~\\
Our main contributions are:
\begin{itemize}
    \item We automate an image generation pipeline that also includes labels, bounding boxes and a quality assessment for the synthetic images.
    \item We show that training on purely synthetic data from our image generation pipeline is sufficient to train an image classifier that can visually predict the car brand on a real photograph.
    \item We describe the required amount of and necessary computation time for synthetic images in our use case.
    \item We include and compare different modes of Stable Diffusion for synthesizing images in our pipeline.
\end{itemize}
\section{Related Work}
Large-scale text-to-image diffusion models can be fine-tuned to augment the ImageNet training set \cite{ImageNet} leading to significant improvements in ImageNet classification accuracy \cite{azizi2023synthetic}. Moreover, the authors of \cite{sariyildiz2023fake} investigate using synthetic images produced with Stable Diffusion \cite{StableDiffusion} when training models for ImageNet classification. Whether and how synthetic images generated from text-to-image generation models can be used for image classification in data-scarce settings and in large-scale model pre-training for transfer learning is considered in \cite{he2022synthetic} using the GLIDE diffusion model \cite{abs-2112-10741}. 

Synthetic data has been successfully used to improve identification and classification tasks in other areas of application such as weather prediction \cite{pub.1147357025}, lung edema identification in chest X-ray images \cite{10508671}, fish species classification \cite{10244720} and the diagnosis of skin diseases \cite{10499667}. In the latter two references Stable Diffusion was used to generate the corresponding synthetic image datasets. Introducing synthetic test data has been proposed as a means to improve model evaluation on diverse and underrepresented population subgroups \cite{05fb0f4e}. 

In the field of vehicle type classification, or, more specifically, car brand and model identification, mainly models trained on real-world data have been investigated so far. Examples are the extension of models trained on limited-size datasets to handle extreme lighting conditions \cite{7868561}, balanced sampling to address the challenge of classifying imbalanced data from visual traffic surveillance sensors \cite{8082506}, improving accuracy of car type classification through the adaptation of specific CNN architecture models \cite{9980771}, as well as adapting deep learning techniques for vehicle color classification \cite{9641933} and vehicle logo recognition \cite{LU2021623}. The detection, recognition, and counting of vehicles based on their car types using YOLOv4  and a combination of YOLOv5 and ResNet has been investigated in \cite{9936874} and \cite{10060631}, respectively. 

With the emergence of state-of-the-art diffusion models the resulting high-quality and diverse synthetic images can be used to enhance detection and classification of different traffic scenarios. Example applications are the detection and classification of armored fighting vehicles \cite{9855933}, parking space classification by utilizing automatically-annotated synthetic video data \cite{8500453}, vehicle detection in aerial and satellite imagery \cite{ORIC2024110105}, the augmentation of virtual objects for car detection in outdoor driving scenarios \cite{Alhaija2018IJCV}, and the classification of different car models by proposing a shallow CNN architecture and a synthetic image augmentation technique  in Keras \cite{9452842}.

One of the challenges in using synthetic data for the improvement of image classification is the domain gap between the synthetic and the real data possibly leading to reduced performance when models trained on one domain are applied to the other \cite{9816205}. In \cite{10208914}, it is shown that performance can be improved significantly by improving the diversity of images in the generated dataset. Furthermore, the authors of \cite{9661221} design a method of generating images to analyze and reduce this domain gap specifically in the application of car detection.
\section{Method} \label{sec:Method}
\begin{figure}[b]
    \centering
    \includegraphics[width=\textwidth]{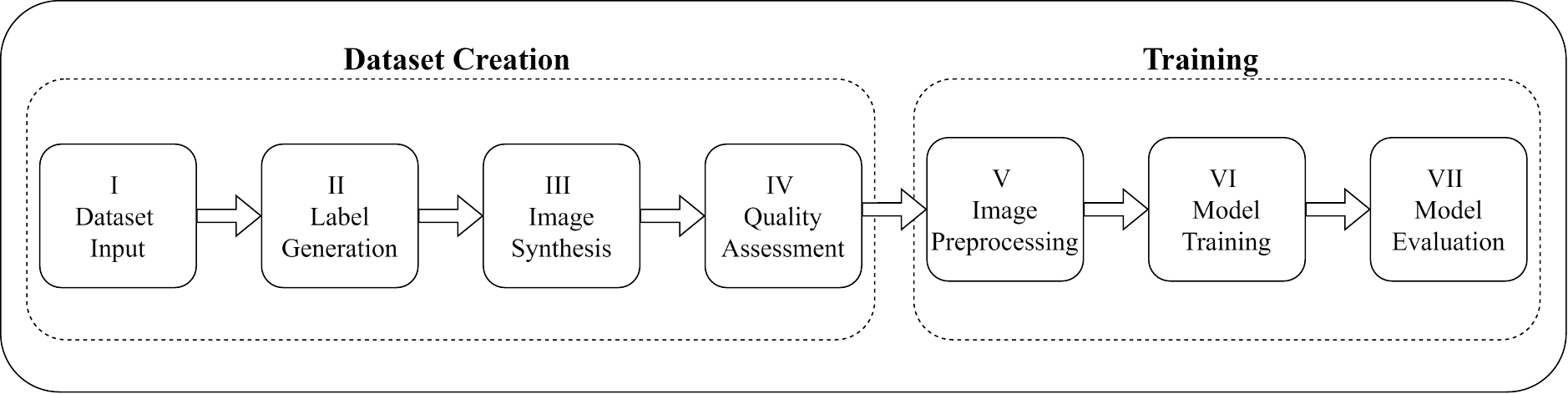}
    \caption{The scheme of the developed pipeline consisting of dataset creation and training. The illustrated pipeline produces a dataset of synthetic images with corresponding labels and bounding boxes. The dataset can be used to automatically train an image classification model.}
    \label{fig:pipeline-figure}
\end{figure}
The goal of this work is to develop a pipeline (illustrated in Figure \ref{fig:pipeline-figure}), which can generate a balanced dataset for a computer vision task with otherwise limited available labeled data. We show that this pipeline is viable in a real-world application to successfully train a machine learning model on it. 

\textbf{Preliminaries} Our proposed automatic pipeline for generating image datasets relies on Stable Diffusion \cite{StableDiffusion} and YOLOv8 \cite{YOLO}. 
Stable Diffusion is a novel and open-source approach for image synthesis. In this research we focus on two modes of Stable Diffusion. 
The mode \textit{Text-to-Image} enables us to create an image from just a text prompt. By altering the prompts and the parameters that control the influence of the prompt we can tweak the resulting images to subjectively fit our use-case.
\textit{Image-to-Image} is another mode that enables us to create an image on base of an already existing image in combination with a text prompt. 
This mode is designed to alter the already existing image whilst preserving the characteristics of the image.
In our experience Stable Diffusion produces undesired results in rare cases. For example, we recorded instances where more than one car was present on the output images.

The images generated by Stable Diffusion do not have an inherent bounding box.
Therefore, we use YOLOv8 in object detection mode to enrich the generated images by their bounding boxes.
YOLOv8 also provides a score for a detected bounding box which can be used to assess the quality of a generated image.

\textbf{I Dataset Input}
The step \textit{Dataset Input} encompasses the steps for retrieving and preparing data for the use case such as car brands, models and build years in our example.
For this work we used the official car registrations \cite{KBA_FZ12,KBA_FZ2,KBA_SV42} from the Federal Motor Transport Authority of Germany (Kraftfahrt-Bundesamt). 
They provide data for registered car models in Germany. 
The features are the vehicle class, the brand, the model name, the build years and the registered number of cars in each category. 
An example for a car model is the Skoda Karoq, a SUV of the brand Skoda produced in the years 2017, 2018 and 2020.
However, we ruled out production years earlier than 1990 as they rarely occur in everyday traffic. 
While it might be possible to generate suitable data even for rare brands, we limit our approach to the brands shown in Figure \ref{fig:Selected Brands}. 
By focusing on these brands, we aim to include the often occurring brands such as Volkswagen and Ford but also more rarely occurring brands such as Skoda and Renault.

\textbf{II Label Generation}
This step is the transformation of the relevant data from the previous step into labels that match our desired distribution. 
In order to balance the dataset, we use a multi-step hierarchical uniform probability distribution. 
In the first hierarchy step each brand has the same probability. 
For each brand the probability of the respective car models is also uniformly distributed. The same principle applies to the construction years for each model.
We aim to avoid biases towards one brand or certain intervals in production years. 
We control the color of a car by also using a uniform distribution over the most common car colors. 

\begin{figure}[t]
    \centering
    \includegraphics[width=.9\textwidth]{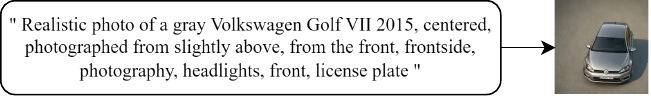}
    \caption{The prompt used to generate the images with Stable Diffusion alongside a generated image using Text-to-Image. The substring \textit{gray Volkswagen Golf VII 2015} is changed accordingly for different car models.}
    \label{fig:prompt}
\end{figure}
\textbf{III Image Synthesis}
In this pipeline step we sample from the labels created in the previous step and create a prompt for each sample as shown in Figure \ref{fig:prompt}. 
The prompt describes the car model in detail as well as the desired perspective. 
The perspective is pretty similar between the different traffic cameras: Shot from the front, from above and centered. 
We then use the respective Stable Diffusion modes to create images using the generated prompts. 

We create two datasets, one for Text-to-Image and one for Image-to-Image.
In both cases we use Stable Diffusion XL Turbo in the respective mode.
If not noted otherwise, we use the standard parameters set by the Python Diffusers library \cite{diffusers}.
For Text-to-Image we use four inference steps with one image per prompt and a guidance scale of zero. This guidance scale is recommended in the documentation for the usage of this model \cite{sdxl-turbo-documentation}.
For Image-to-Image we use ten inference step with a guidance scale of 0.4 and a strength of 0.6.
These parameters are manually tuned to subjectively fit the desired output.

\begin{figure}
\centering

    \begin{tabular}{cccccc}
    \includegraphics[width=0.15\textwidth,height=0.15\textwidth]{fig_brand_volkswagen.jpg}
    & \includegraphics[width=0.15\textwidth,height=0.15\textwidth]{fig_brand_ford.jpg} \hspace{2pt}
    & \includegraphics[width=0.15\textwidth,height=0.15\textwidth]{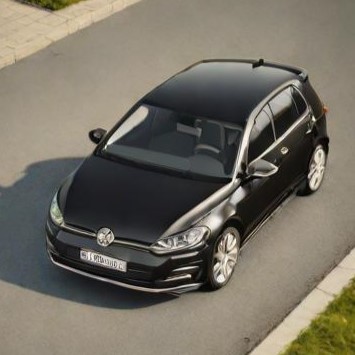} 
    & \includegraphics[width=0.15\textwidth,height=0.15\textwidth]{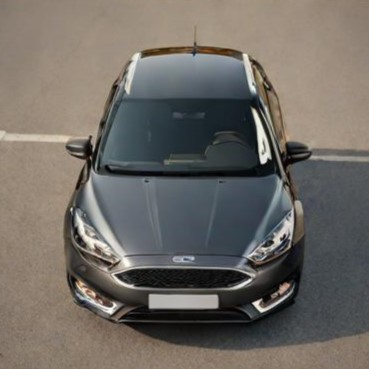} \hspace{2pt}
    & \includegraphics[width=0.15\textwidth,height=0.15\textwidth]{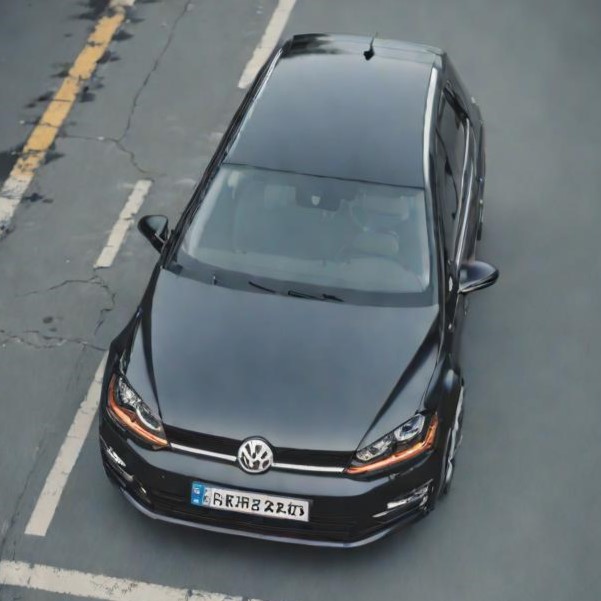}
    & \includegraphics[width=0.15\textwidth,height=0.15\textwidth]{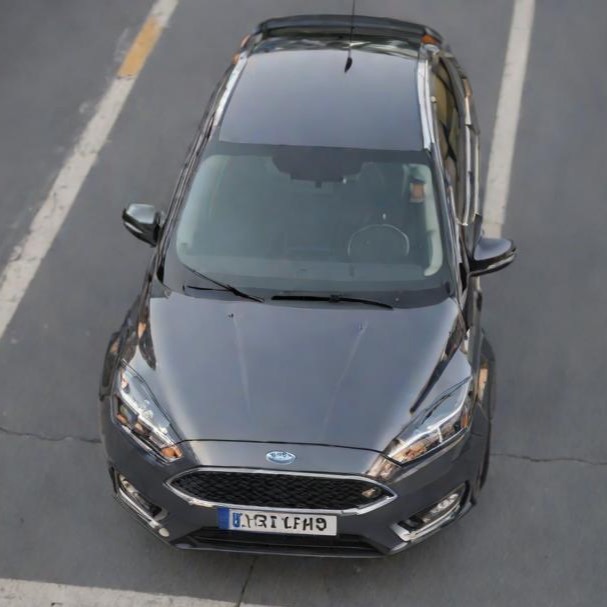}  \\     
    \multicolumn{2}{c}{Real Photographs} & \multicolumn{2}{c}{Text-to-Image} & \multicolumn{2}{c}{Image-to-Image} \\
    \end{tabular}

    \caption{Illustration of differences between real images and modes of image generation. Text-to-Image tends to encompass more perspectives contrary to the narrow range of perspectives with Image-to-Image.}
    \label{fig:Pipeline Image Results}
\end{figure}

When using Stable Diffusion in Image-to-Image mode we also have to provide a base image in conjunction with a prompt as the input for the model.
To create these base images we use real photographs of cars at different positions on the road cropped to the car with padding. 
We use photographs of cars with brands we do not want to classify as they are not present within the validation and test dataset. 
With these base images we intend to implicitly give the desired perspective so that the generated images strongly resemble the real images.
The input base image for this mode is scaled up to 720x720 pixels as this is the minimal size for Stable Diffusion XL.
Figure \ref{fig:Pipeline Image Results} illustrates real photographs compared to images generated by Text-to-Image and Image-to-Image. 

\setlength{\tabcolsep}{1pt}

\setlength{\tabcolsep}{6pt}

\begin{figure}[b]
\centering
    \begin{tabular}{cccc}
        \includegraphics[width=0.2\textwidth]{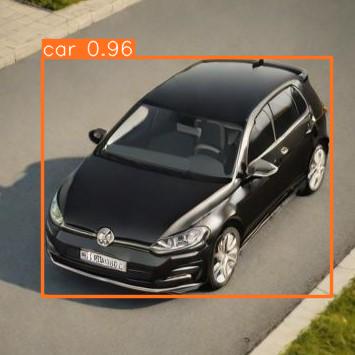}
        & \includegraphics[width=0.2\textwidth]{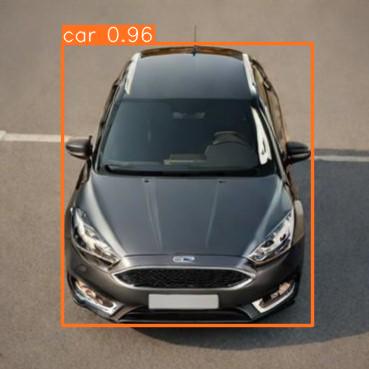}
        & \includegraphics[width=0.2\textwidth]{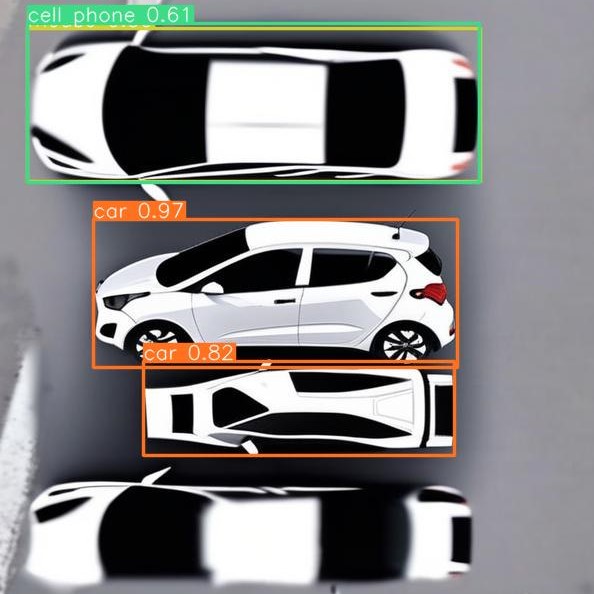}
        & \includegraphics[width=0.2\textwidth]{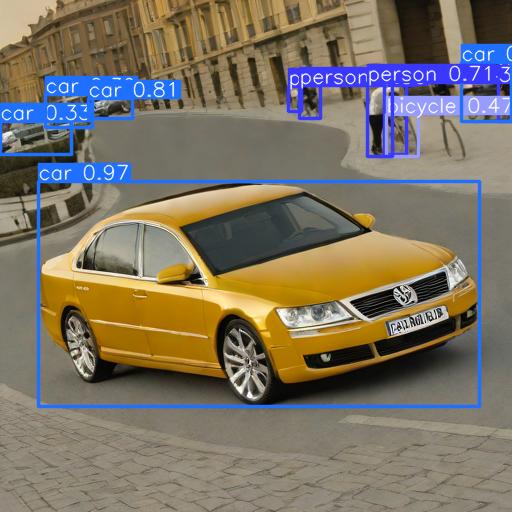}
    \end{tabular}
    \caption{Bounding Boxes detected with YOLOv8x. This allows to crop the image and provides a confidence score for the presence of a car. It also allows to automatically sort out the two undesired images on the right where more than one car is detected.}
    \label{fig:YOLO Bounding Boxes}
\end{figure}

\textbf{IV Quality Assessment}
The output of image synthesis models such as Stable Diffusion normally matches the expected output. In most of the cases there is exactly one car as the main subject of the image. The location and the size of the image's subject differs. Therefore an object detection model such as YOLOv8x has to be used to automatically determine the bounding boxes (see Figure \ref{fig:YOLO Bounding Boxes}).
\setlength{\tabcolsep}{2pt}
\setlength{\tabcolsep}{6pt}
However, we observe that in rare cases Stable Diffusion produces something other than the desired output of a singular car (see Figure \ref{fig:YOLO Bounding Boxes}). 
Therefore, we use YOLOv8x in object detection mode on each image. 
This allows us to automatically confirm that there is exactly one car in the image. 
It further assesses the quality of the generated image as YOLO provides a score for the certainty of a detected bounding box. 
We then crop the image to the bounding boxes of the detected car.
This allows the automatic generation of the bounding boxes for the dataset as well as increasing the quality by sorting out unwanted images.

\textbf{V Image Preprocessing}
Stable Diffusion models allow to specify the dimensions of the resulting image. However, the subject of the images varies in size so that the images cropped to their subjects' bounding boxes differ in aspect ratios and sizes. 
We transformed the images to 64x64 pixels. 
The small resolution is chosen as the traffic cameras in our use case record in HD and cars at different positions in the image may therefore have a similar resolution.
Random Rotation as a classical data augmentation method is also applied to the dataset.

\textbf{VI Model Training}
We then use the generated and pre-processed image dataset to train image classifiers. 
In using the model Resnet-18 that is pre-trained on ImageNet \cite{ImageNet}, we apply the principle of transfer learning \cite{TransferLearning,TransferLearningExample}. 
For adapting this model we replace the last fully connected layer by a new fully connected layer with the same input size and an output size of eight which encodes the classes we want to classify.
We do not lock any pretrained layer.

\textbf{VII Evaluation}
The performance of the model is validated against real-world data recorded by traffic cameras mounted in Lemgo, Germany. 
These images were manually labeled. Images of the same car at different positions can exist in the datasets as the cars are moving forward. 
The split between validation and test dataset is therefore performed based on location of the respective camera and the time of recording. 
Due to the biases in the real world and the described split, the classes are unevenly distributed. 
\section{Experiments and Results}
The time for generating images with Stable Diffusion depends on output size and the number of inference steps. 
With Stable Diffusion XL Turbo the inference steps can be kept lower than for the normal Stable Diffusion XL. 
For this model the output size should be bigger than 720 pixels in both dimensions.
In the following we consider the performance for the parameters described in Section \ref{sec:Method} (Method) when run on a Nvidia RTX 3060.
When using Text-to-Image (four inference steps) it took $0.85$ seconds on average to produce one image.
For Image-to-Image (ten inference steps) this took $2.33$ seconds. 
These durations account for image synthesis, bounding box detection, automatic quality assessment and storing of the image.
This performance is remarkable as it provides a fast and automatic way to generate usable datasets within days or even hours. This process can be sped up by using further optimizations of the model, faster hardware or possibly less inference steps.

We retrain Resnet-18 on varying datasets. The model is trained on the images in random order with an exponentially decaying learning rate starting at $0.01$ and the stochastic gradient descent optimizer. An epoch for training the Resnet-18 on 100.000 images takes approximately 18 seconds on a Nvidia RTX 3060.

We evaluate the performance of this model regarding the different modes of image synthesis and the required amount of data.  
Our approach enables us to control the distribution of the generated training datasets to be perfectly balanced.
With this work we evaluate the feasibility of synthetic images as training data for models that work on real world data.
As we use real world photographs of cars in traffic in Lemgo, Germany, we have an unbalanced dataset. The split, described earlier, results in a validation dataset consisting of 1503 images and a test dataset consisting of 1317 images (in detail in Table \ref{table:validation_datasets}).

\begin{table}
\caption{Distribution of brands in validation and test dataset. We can see that the data is unbalanced and strongly biased towards Volkswagen.}
~\\
\centering
\begin{tabular}{|l|l|l|}
\hline
Brand        & Validation Dataset & Test Dataset \\ \hline
Volkswagen   & 533                & 443          \\ \hline
Ford         & 162                & 143          \\ \hline
BMW          & 171                & 146          \\ \hline
Audi         & 130                & 142          \\ \hline
Opel         & 133                & 159          \\ \hline
Mercedes     & 167                & 123          \\ \hline
Renault      & 91                 & 74           \\ \hline
Skoda        & 116                & 87           \\ \hline
\end{tabular}
\label{table:validation_datasets}
\end{table}

\begin{table}[!b]
\caption{Confusion matrix for Resnet-18 trained on the mixed dataset of 400,000 images. The characteristics of this confusion matrix are typical for the results on this dataset.}
~\\
\centering
\begin{tabular}{|c|lllllllll|}
\hline
\multicolumn{1}{|l|}{}  & \multicolumn{9}{c|}{Predicted} \\ \hline
\multirow{9}{*}{Actual} & \multicolumn{1}{l|}{}           & \multicolumn{1}{l|}{Volkswagen}   & \multicolumn{1}{l|}{Ford}          & \multicolumn{1}{l|}{BMW}          & \multicolumn{1}{l|}{Audi}          & \multicolumn{1}{l|}{Opel}          & \multicolumn{1}{l|}{Mercedes}      & \multicolumn{1}{l|}{Renault}       & Skoda         \\ \cline{2-10} 
                        & \multicolumn{1}{l|}{Volkswagen} & \multicolumn{1}{l|}{\textbf{0.9}} & \multicolumn{1}{l|}{0.03}          & \multicolumn{1}{l|}{0.01}         & \multicolumn{1}{l|}{}              & \multicolumn{1}{l|}{}              & \multicolumn{1}{l|}{0.03}          & \multicolumn{1}{l|}{0.02}          &               \\ \cline{2-10} 
                        & \multicolumn{1}{l|}{Ford}       & \multicolumn{1}{l|}{0.13}         & \multicolumn{1}{l|}{\textbf{0.81}} & \multicolumn{1}{l|}{}             & \multicolumn{1}{l|}{}              & \multicolumn{1}{l|}{0.02}          & \multicolumn{1}{l|}{0.04}          & \multicolumn{1}{l|}{}              &               \\ \cline{2-10} 
                        & \multicolumn{1}{l|}{BMW}        & \multicolumn{1}{l|}{0.05}         & \multicolumn{1}{l|}{0.06}          & \multicolumn{1}{l|}{\textbf{0.7}} & \multicolumn{1}{l|}{0.01}          & \multicolumn{1}{l|}{0.01}          & \multicolumn{1}{l|}{0.13}          & \multicolumn{1}{l|}{0.03}          & 0.01          \\ \cline{2-10} 
                        & \multicolumn{1}{l|}{Audi}       & \multicolumn{1}{l|}{0.08}         & \multicolumn{1}{l|}{0.13}          & \multicolumn{1}{l|}{}             & \multicolumn{1}{l|}{\textbf{0.73}} & \multicolumn{1}{l|}{0.01}          & \multicolumn{1}{l|}{0.04}          & \multicolumn{1}{l|}{}              & 0.01          \\ \cline{2-10} 
                        & \multicolumn{1}{l|}{Opel}       & \multicolumn{1}{l|}{0.14}         & \multicolumn{1}{l|}{0.14}          & \multicolumn{1}{l|}{0.01}         & \multicolumn{1}{l|}{}              & \multicolumn{1}{l|}{\textbf{0.55}} & \multicolumn{1}{l|}{0.11}          & \multicolumn{1}{l|}{0.03}          & 0.01          \\ \cline{2-10} 
                        & \multicolumn{1}{l|}{Mercedes}   & \multicolumn{1}{l|}{0.07}         & \multicolumn{1}{l|}{0.03}          & \multicolumn{1}{l|}{0.01}         & \multicolumn{1}{l|}{0.02}          & \multicolumn{1}{l|}{0.04}          & \multicolumn{1}{l|}{\textbf{0.81}} & \multicolumn{1}{l|}{0.02}          & 0.01          \\ \cline{2-10} 
                        & \multicolumn{1}{l|}{Renault}    & \multicolumn{1}{l|}{0.32}         & \multicolumn{1}{l|}{}              & \multicolumn{1}{l|}{0.07}         & \multicolumn{1}{l|}{}              & \multicolumn{1}{l|}{0.07}          & \multicolumn{1}{l|}{0.03}          & \multicolumn{1}{l|}{\textbf{0.51}} &               \\ \cline{2-10} 
                        & \multicolumn{1}{l|}{Skoda}      & \multicolumn{1}{l|}{0.15}         & \multicolumn{1}{l|}{0.06}          & \multicolumn{1}{l|}{0.07}         & \multicolumn{1}{l|}{0.01}          & \multicolumn{1}{l|}{0.07}          & \multicolumn{1}{l|}{0.05}          & \multicolumn{1}{l|}{0.03}          & \textbf{0.56} \\ \hline
\end{tabular}

\label{table:Cm1}
\end{table}

\begin{figure}
\begin{tikzpicture}
    \centering
    \begin{axis}[
        width=0.95\textwidth,
        height=0.45\textwidth,
        xlabel=Dataset Size,
        ylabel=Accuracy,
        xmode=log,
        xmin=10000, xmax=450000,
        xtick={12000, 25000, 50000, 100000, 200000, 400000},
        xticklabels={12K, 25K, 50K, 100K, 200K, 400K},
        ymin=0.48, ymax=0.82,
        ytick={0.5, 0.6, 0.7, 0.8},
        grid=both,
        major grid style={line width=.2pt,draw=gray!50},
    ]
        \addplot[smooth,color=orange,mark=square*] plot coordinates {
            (12000, 0.61)
            (25000, 0.66)
            (50000, 0.67)
            (100000, 0.7)
            (200000, 0.7)
            (400000, 0.71)
            };
        \addlegendentry{Combined}
        \addplot[smooth, opacity=0.3, name path=TTI1,color=orange,forget plot] plot coordinates {
            (12000, 0.58)
            (25000, 0.64)
            (50000, 0.66)
            (100000, 0.69)
            (200000, 0.7)
            (400000, 0.7)
        };
        \addplot[smooth, opacity=0.3, name path=TTI2,color=orange,forget plot] plot coordinates {
            (12000, 0.64)
            (25000, 0.68)
            (50000, 0.68)
            (100000, 0.71)
            (200000, 0.7)
            (400000, 0.72)
        };
        \addplot[color=orange, opacity=0.4, fill opacity=0.1, forget plot] fill between[of=TTI1 and TTI2];

        \addplot[smooth,mark=triangle*,color=red] plot coordinates {
            (12000, 0.54)
            (25000, 0.58)
            (50000, 0.61)
            (100000, 0.62)
            (200000, 0.63)
            (400000, 0.63)
        };
        \addlegendentry{Image-to-Image}
        \addplot[smooth, opacity=0.3, name path=TTI1,color=red,forget plot] plot coordinates {
            (12000, 0.49)
            (25000, 0.56)
            (50000, 0.59)
            (100000, 0.60)
            (200000, 0.61)
            (400000, 0.60)
        };
        \addplot[smooth, opacity=0.3, name path=TTI2,color=red,forget plot] plot coordinates {
            (12000, 0.59)
            (25000, 0.60)
            (50000, 0.63)
            (100000, 0.64)
            (200000, 0.65)
            (400000, 0.66)
        };
        \addplot[color=red, opacity=0.4, fill opacity=0.1, forget plot] fill between[of=TTI1 and TTI2];
        
        \addplot[smooth,color=blue,mark=*] plot coordinates {
            (12000, 0.56)
            (25000, 0.61)
            (50000, 0.60)
            (100000, 0.62)
            (200000, 0.62)
            (400000, 0.64)
        };
        \addlegendentry{Text-to-Image}
        \addplot[smooth, opacity=0.3, name path=TTI1,color=blue,forget plot] plot coordinates {
            (12000, 0.58)
            (25000, 0.63)
            (50000, 0.62)
            (100000, 0.63)
            (200000, 0.63)
            (400000, 0.65)
        };
        \addplot[smooth, opacity=0.3, name path=TTI2,color=blue,forget plot] plot coordinates {
            (12000, 0.54)
            (25000, 0.59)
            (50000, 0.58)
            (100000, 0.61)
            (200000, 0.61)
            (400000, 0.63)
        };
        \addplot[color=blue, opacity=0.4, fill opacity=0.1, forget plot] fill between[of=TTI1 and TTI2];
        
    \end{axis}
\end{tikzpicture}
\caption{Training Results for Resnet-18 trained on the different dataset sizes. Model performance and dataset size correlate. The experiments were performed five times per dataset size and Stable Diffusion mode. The graph shows a confidence interval of one standard deviation.}
    \label{fig:Training Results}
\end{figure}
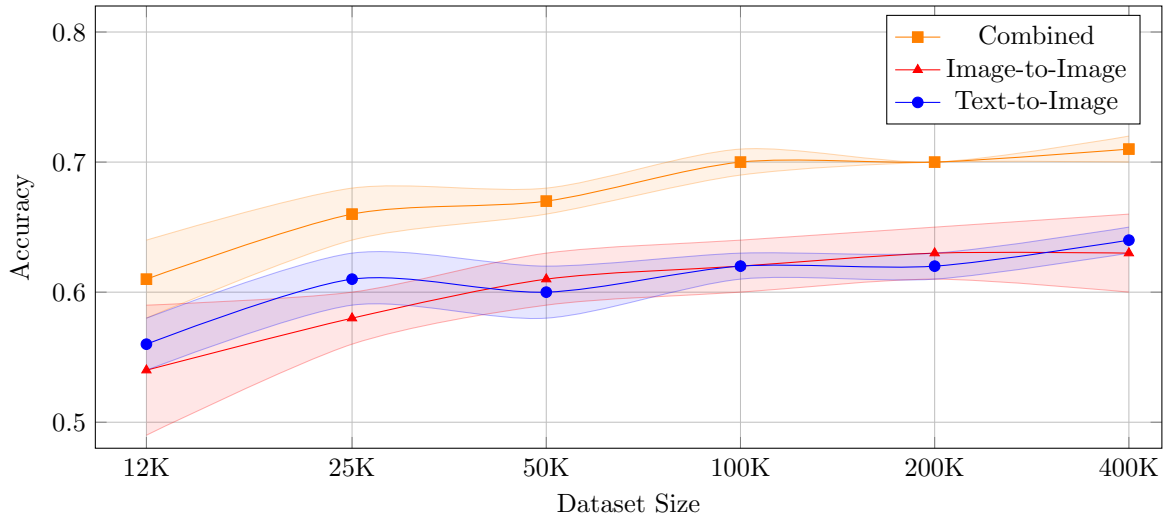

Figure \ref{fig:Training Results} shows the training results for different dataset sizes and datasets generated by Image-to-Image mode, Text-to-Image mode and a combination of these modes. These results show that we are able to train a model on the generated images that can exceed the primitive baseline by far, even reaching up to 75\% in accuracy on the given eight classes.

We observe that the performance of the retrained Resnet-18 model has a reliable and stable performance on both modes of image generation and both modes combined. 
The performance of this model in regards to the mode of image generation does not differ significantly. On Text-to-Image the model shows a slightly more stable performance.
The Resnet-18 model clearly benefits from using images from both modes combined leading to a score of 75\% at maximum with a dataset size of 400.000 images. We can see that the model performance correlates with the dataset size and the variety in images, as the performance when trained on images from both modes combined surpasses the performance when trained only on images from one mode. For example: A model trained on 50,000 images from both modes surpasses a model trained on 400,000 images from a singular source.

The typical confusion matrix given in Table \ref{table:Cm1} shows that the retrained models are typically biased towards predicting Volkswagen which is the most common brand in the photographs of the real traffic. We also observe a difference in accuracy per brand. 
Volkswagen, Ford, BMW, Audi and Mercedes achieve a performance of at least 70\% while 
Opel, Renault and Skoda are only correct in about half of the cases.
\section{Conclusion}
We are able to train image classifiers for real world data solely on synthetic images that require no human labeling. 
The images we evaluate the classifiers on are taken in real moving traffic. 
Therefore, we face challenges such as a large variety of objects, image artifacts, different lighting, low resolution and motion blur. 
The generated data is sufficient to exceed the primitive baseline by far.
These results are achieved whilst needing human work only for engineering the pipeline, tweaking hyperparameters and labeling the validation and test images.
Thus the engineered pipeline may illustrate a potential approach to overcome challenges associated with traditional data acquisition methods.

On average the Resnet-18 performs better when retrained on the combined images instead of the same amount from just one mode of image generation. 
This may result from the fact that both modes combined cover a broader variety regarding the characteristics of the images. 
We assume that using varying prompts and varying parameters for the Stable Diffusion model could increase the variation in images and could therefore be beneficial.

To illustrate one possibility of this pipeline: With our pipeline we are able to create a perfectly balanced dataset of 100.000 images by using both modes combined.
We can directly train a Resnet-18 model on this generated dataset. The time to perform this consecutively on a singular Nvidia RTX 3060 without further optimizations sums up to about two days. This provides a solid baseline model on short term with very little human work required.

There are significant differences in performance of the retrained Resnet-18 per class (see Table \ref{table:Cm1}). 
These differences may be explained by biases inside of the Stable Diffusion models as they are more likely trained on more images of Volkswagen than images of Skoda. 
Another may be that, on one hand, brands like Volkswagen, Mercedes, BMW and Audi have very prominent visual features that are easy recognizable for humans.
On the other hand, earlier models of Renault have very small logos and Skoda has a dark radiator grill with only a small logo as an identifier.
This also can attribute to a lower performance for these brands.

The pipeline introduced in this work is possible as we can automatically assess the output of the Stable Diffusion model with YOLO which also allows us to crop the image to the relevant parts (see Figure \ref{fig:pipeline-figure} Step IV).
For other computer vision tasks with classes that are a subset of the classes YOLO can predict, we can adapt the pipeline easily.
However, for completely other classes one would have to engineer another way to implement the fourth step of the pipeline.
As this presents a challenge, the possible use cases of the introduced pipeline are limited by the capabilities of object detection models like YOLO.
Another limitation lies within the capability of Stable Diffusion as it is unlikely that these models can generate usable images for every situation.
\section{Acknowledgments}
The authors express their sincere gratitude to the Fraunhofer IOSB-INA for providing the computational resources for this project.

Markus Lange-Hegermann acknowledges support from the German Federal Ministry of Education and Research (BMBF) for the project SyDaPro with grant number 01IS21066A.

\newpage

\printbibliography

\end{document}